\def\year{2022}\relax
\documentclass[letterpaper]{article} 
\usepackage{aaai22}  
\usepackage{times}  
\usepackage{helvet}  
\usepackage{courier}  
\usepackage[hyphens]{url}  
\usepackage{graphicx} 
\usepackage{natbib}  
\usepackage{caption} 
\DeclareCaptionStyle{ruled}{labelfont=normalfont,labelsep=colon,strut=off} 
\usepackage{algorithm}
\usepackage{algorithmic}
\usepackage{amsmath}
\usepackage{amsfonts}
\usepackage{booktabs}
\usepackage{makecell}
\usepackage{hyperref}
\usepackage{url}
\usepackage{changes}
\usepackage{subfigure}

\usepackage{afterpage}
\usepackage{enumitem}

\usepackage{newfloat}
\usepackage{listings}
\floatstyle{ruled}
\newfloat{listing}{tb}{lst}{}
\floatname{listing}{Listing}
\title{Where Did the President Visit Last Week? Detecting Celebrity Trips from News Articles}

\author{
    Kai Peng\equalcontrib, Ying Zhang\equalcontrib, Shuai Ling, Zhaoru Ke, Haipeng Zhang\thanks{Corresponding author.}
}
\affiliations{
    ShanghaiTech University\\

    \{pengkai, zhangying12022, lingshuai, kezhr, zhanghp\}@shanghaitech.edu.cn
}

\usepackage{bibentry}

\begin{document}

\maketitle

\begin{abstract}

Celebrities’ whereabouts are of pervasive importance. For instance, where politicians go, how often they visit, and who they meet, come with profound geopolitical and economic implications. Although news articles contain travel information of celebrities, it is not possible to perform large-scale and network-wise analysis due to the lack of automatic itinerary detection tools. To design such tools, we have to overcome difficulties from the heterogeneity among news articles:
1) One single article can be noisy, with irrelevant people and locations, especially when the articles are long.
2) Though it may be helpful if we consider multiple articles together to determine a particular trip, the key semantics are still scattered across different articles intertwined with various noises, making it hard to aggregate them effectively.
3) Over 20\% of the articles refer to  celebrity trips indirectly, instead of using the exact celebrity names or location names, leading to large portions of trips escaping regular detecting algorithms.
We model text content across articles related to each candidate location as a graph to better associate essential information and cancel out the noises. Besides, we design a special pooling layer based on attention mechanism and node similarity, reducing irrelevant information from longer articles. To make up the missing information resulted from indirect mentions, we construct knowledge sub-graphs for named entities (person, organization, facility, etc.). Specifically, we dynamically update embeddings of event entities like the G7 summit from news descriptions since the properties (date and location) of the event change each time, which is not captured by pre-trained event representations. The proposed CeleTrip jointly trains these modules, which outperforms all baseline models and achieves 82.53\% in the F1 metric.
By open-sourcing the first tool and a carefully curated dataset for such a new task, we hope to facilitate relevant research in celebrity itinerary mining as well as the social and political analysis built upon the extracted trips.
\end{abstract}

\section{Introduction}
\label{intro}
Celebrity itineraries have received extensive attention in social science, with applications to politics~\cite{brace1993presidential,goldsmith2021does}, economics~\cite{boianovsky20182017}, arts~\cite{badham2017social}, and cultural history~\cite{schich2014network}. Specifically, researchers analyze both domestic and international situations using the travels of politicians. \citet{ostrander2019presidents} point out that the president is more likely to travel overseas when there is a split government. Furthermore, itineraries of artists also serve the government's decision-making. For example, based on the mobility of artists, \citet{toplak2017artists} demonstrates that administrative barriers have impeded population mobility within the EU compared to the past.

Despite its value, the scarcity of available trip data prevents relevant studies from being extended to large-scale and network-wise. 
Though official websites may be an authoritative and accurate source for such studies~\cite{ostrander2019presidents}, they are infrequently updated and the coverage is limited.
For instance, the Japanese Ministry of Foreign Affairs\footnote{\url{https://www.mofa.go.jp/mofaj/gaiko/bluebook/index.html}} releases the diplomatic blue book only once a year. Therefore, it is necessary to more efficiently obtain celebrity itineraries.

\begin{figure}[!htb]
    \centering
    \includegraphics[width=\linewidth]{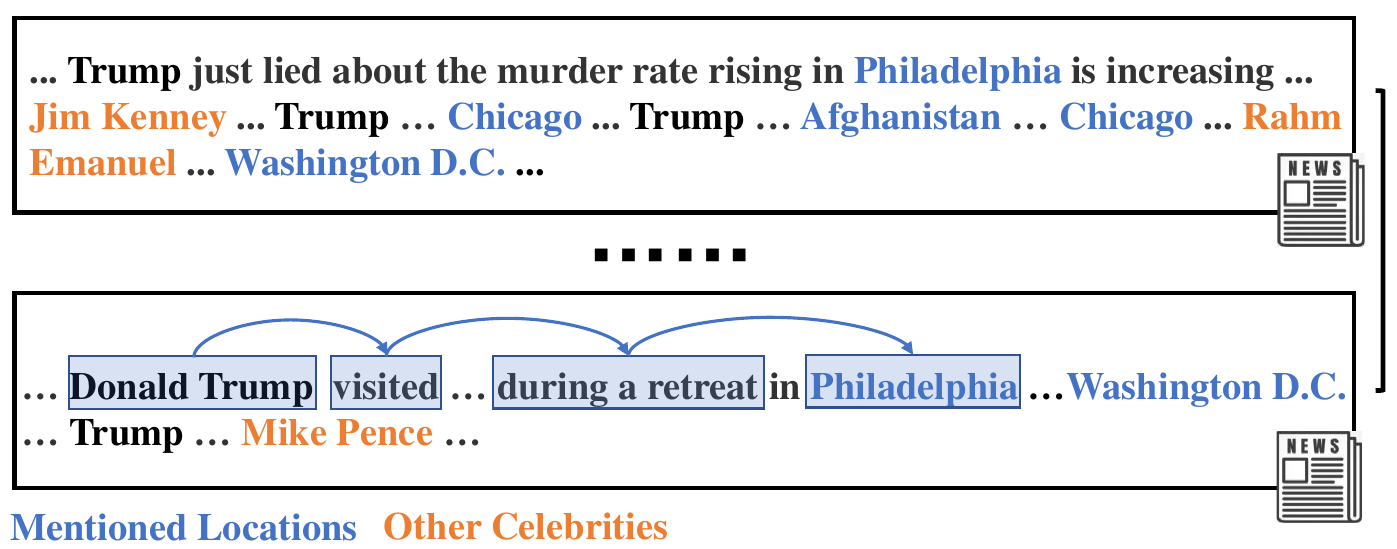}
    \caption{Example of extracting trip information of Donald Trump from news articles on 2017-01-26.}
    \label{fig:intro_case}
\end{figure}

Online news contains a wealth of information related to celebrities, including their trips~\cite{piskorski2020timelines}, thereby serving as a convenient source for acquiring copious celebrity itineraries. 
Our task is to find the time and locations of the trips made by given celebrities, from news articles. A reasonable intuition would be first narrowing down the search space to the articles containing the target celebrity's name, associated with the target date, further recognizing place names with off-the-shelf Named Entity Recognition (NER) tools~\cite{lingad2013location,li2020survey} as candidate locations, and choosing the most frequent one as the visited place.
However, as shown in Figure~\ref{fig:intro_case}, interfering information, including locations visited by other celebrities, always scatters around the travel destination, potentially impeding the task. According to our statistics, more than 50\% of the locations mentioned in the news are not actually visited by the target celebrities. Even after sorting the locations by frequency, the results remain unfavorable (see Experimental Results). 
As an important refinement, the textual context of locations should be considered to distinguish travel locations from others.
Figure~\ref{fig:intro_case} again demonstrates an intuitive example -- words like \textit{visited} (in the bottom article) signal the presence of Donald Trump in \textit{Philadelphia}, while \textit{Washington D.C.} without such words seems a less likely destination.

Besides the current task, textual descriptions are also utilized in solving detection problems, with the most relevant one being Person-Related Event Detection (PRED). Instead of mining public news reports on celebrities, PRED aims to detect predefined personal life events from social media by classifying user-posted text, especially tweets, into categories such as having children~\cite{dickinson2015identifying} or visiting~\cite{yen2021personal}. 
Though we can also model our task by classifying the aforementioned candidate locations into positives (actually visited) and negatives (actually not visited) given their textual contexts, adopting the PRED methods can be problematic.
First, PRED methods classify one (short) tweet at a time~\cite{yen2019personal} and they become less effective when dealing with multiple long articles. In our scenario, the mentions of celebrities and locations are scattered in much longer text, usually across different articles. Regarding the example in Figure~\ref{fig:intro_case}, we need sequential combinations of \textit{Trump}, \textit{visit}, and \textit{Philadelphia} that are far apart, as well as information about other candidate locations from various articles, to infer Trump's itinerary for that day. Traditional bag-of-words based PRED approaches~\cite{akbari2016tweets} ignore the words' sequence and associations, while methods that build upon convolutional neural networks (CNNs)~\cite{nguyen2017robust} and recurrent neural networks (RNNs)~\cite{yen2018detecting} focus on local semantics and short sequence~\cite{peng2019fine}, overlooking long-range cross-document dependency. To incorporate all these, representing each candidate location in the binary classification problem with a graph of its contextual words, the documents mentioning it, and their associations, may be a more suitable choice.


The second major difference between our task and PRED is that we need to deal with more irrelevant celebrities and locations, since a single news article is much longer and the noises multiply when multiple articles are considered. The PRED task assumes the events are related to the specific user who posts the tweet, without delving into the identities of event participants~\cite{wu2019event,yen2019personal}. While in our scenario, we have to avoid confusing itineraries of other celebrities with the one of the target celebrity, by focusing on measuring how relevant each candidate location is to the target celebrity on its corresponding graph.

Another challenge faced in our scenario is the so-called ``implicit trips,'' in which the target celebrities and/or their visited locations are not directly mentioned by names, and these cases account for nearly 20\% of all trips in our dataset. For example, the sentence ``...Taylor Swift did something bad for the opening of the 2018 American Music Awards...from the Microsoft Theater...'' does not directly reveal Taylor's geographic location despite indicating her appearance at an event (2018 American Music Awards) and a particular entity (Microsoft Theater). 
Besides, for another common kind of implicit trips, news articles refer to politicians by their job titles, instead of their names. Background information of celebrities and other well-known entities can always be found in knowledge bases such as Wikidata~\cite{vrandevcic2014wikidata}, while for specific events that are often absent in the knowledge bases, news articles published around the target date that cover these events may be a better source of information, especially when mining the latest itineraries.

\begin{figure*}[!htb]
    \centering
    \includegraphics[width=\linewidth]{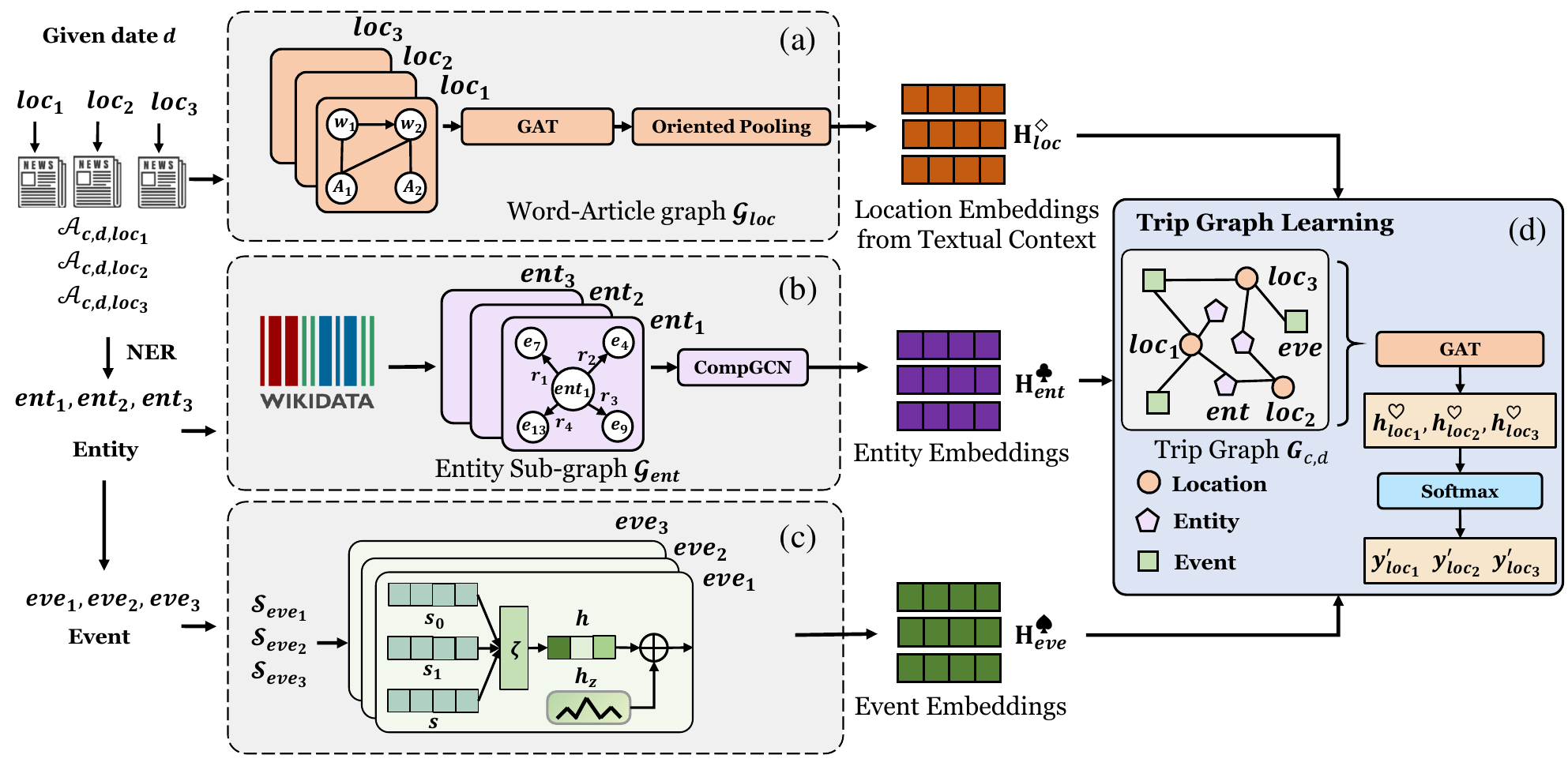}
    \caption{Our framework learns the overall representations of candidate locations and classifies them in \textbf{Trip Graph}, where the textual description of each candidate location is incorporated by the \textbf{Location Embedding Learning}, the knowledge of related entities is supplied by \textbf{Entity Embedding Learning}, and the information of related events is obtained by \textbf{Event Embedding Learning}.}
    \label{fig:framework}
\end{figure*}

In this paper, we present \textbf{CeleTrip}, the first data-driven framework for automatically detecting celebrity itineraries from news articles. Before CeleTrip does its job, we find news articles related to the target celebrity and the particular date, and extract candidate locations from them. CeleTrip (see Figure~\ref{fig:framework}) will determine the actually visited locations through binary classification. As mentioned before, utilizing the textual context of locations poses a challenge of integrating semantics that are far apart. Previous work shows that graph neural networks can capture long-distance dependencies~\cite{sahu2019inter} and help information aggregation through message-passing~\cite{kipf2016semi}. To this end, we construct a \textbf{Word-Article graph} for each location ((a) in Figure~\ref{fig:framework}), which takes its corresponding articles and words in them as nodes to encode the location's context. To uncover the ``implicit trips,'' external sources of knowledge -- Wikidata and news articles, need to be included. Due to their obvious heterogeneity, we decide not to mix them directly with the Word-Article graphs. Instead, we use an \textbf{Entity Sub-graph} for each named entity in the candidate locations' relevant news articles to embed knowledge from Wikidata ((b) in Figure~\ref{fig:framework}), while event-relevant information is represented by event embeddings learned from event-related sentences in news articles ((c) in Figure~\ref{fig:framework}). The resulting entity embeddings and event embeddings, together with these location embeddings extracted from Word-Article graphs, are used to represent corresponding nodes in the \textbf{Trip Graph} of candidate locations, and their relevant entities and events that depict the target celebrity's possible trips on a particular day ((d) in Figure~\ref{fig:framework}). In this way, CeleTrip can jointly model the context of locations with external knowledge from a comprehensive perspective. Moreover, we propose an Oriented Pooling combining node similarity to capture the semantic relation between the candidate location and the target celebrity, reducing the distraction from irrelevant celebrities in the Word-Article graph, as opposed to the regular max pooling in graph convolutional networks (GCNs)~\cite{deng2019learning}.

We summarize our contributions as follows.
\begin{itemize}
    \item We propose a new yet important task of celebrity itinerary detection, construct the largest dataset known so far, and develop a first framework, \textbf{CeleTrip}, for this task. \textbf{CeleTrip} mines news articles and outperforms all baselines with an F1 score of 82.53\% on the dataset.      
    \item We design Word-Article graphs to capture trip-related information scattered among different articles and propose a new graph pooling technique to refine the graph and focus on nodes of interest.
    \item We introduce a simple but effective method to update the representations of event entities based on related news and the attention mechanism.
    \item To facilitate relevant research, we release our dataset that includes 5,687 trips. Besides, we open-source our framework, and the assisting time/location extraction tools\footnote{\url{https://github.com/ZhangDataLab/CeleTrip}}.

\end{itemize}







\section{Problem Statement}

Before we formalize the core problem, we describe the heuristic which is briefly mentioned in Introduction. Since we want to find the locations that celebrity $c$ has visited on date $d$ from news articles, we first narrow down our search to the articles published on date $d$ or containing $d$ in the text, and select these mentioning $c$'s name to be $\mathcal{A}_{c,d}$, which are the articles related to $c$ and $d$. From these articles, we extract all mentioned locations as a set of candidate locations, denoted as $Loc_{c,d}$. The task now becomes: for each candidate location $loc$ in $Loc_{c,d}$, decide whether it is actually visited by celebrity $c$ on date $d$, given articles $\mathcal{A}_{c,d}$ related to $c$ and $d$. 

We train a model $f$ with trainable parameters $\Theta$ to classify $loc\in Loc_{c,d}$ into binary category $y \in \left\{0,1\right\}$, with $y=1$ indicating that $c$ has visited $loc$ on date $d$, and $y=0$ being the opposite:
\begin{align}
    f : \left\{\mathcal{A}, loc, c, d, \Theta\right\} \rightarrow y,
\end{align}
As a side note, to better assist the core task, we develop our own tools for extracting time and locations, as described in the last subsection of Method.

\section{Method}

We elaborate \textbf{CeleTrip} (see Figure~\ref{fig:framework}) in this section. As described in Introduction and Problem Statement, for any target celebrity $c$ on date $d$, the framework learns the overall representations ($\mathbf{h^{\heartsuit}_{loc_1}}$, $\mathbf{h^{\heartsuit}_{loc_2}}$, $\mathbf{h^{\heartsuit}_{loc_3}}$) of the candidate locations and classifies them in (d)~\textbf{Trip Graph Learning}, by incorporating location embeddings from textual descriptions obtained in (a)~\textbf{Location Embedding Learning}, embeddings of the locations' related entities from a knowledge base supplied by (b)~\textbf{Entity Embedding Learning}, and the event embeddings for these event entities  provided by (c)~\textbf{Event Embedding Learning}. The Trip Graph in (d) models $c$'s possible trips on date $d$ -- candidate locations are connected to certain types of co-occurring entities (e.g. person, facility, and event) detected by a Named Entity Recognition (NER) tool and for these event entities, they are represented as event nodes instead, which augment event information from relevant news articles. In the Trip Graph, candidate locations can be indirectly linked if they share the same related entities/events.
Additionally, we describe our tools for extracting time and locations at the end of this section.


\subsection{Location Embedding Learning}
\subsubsection{Word-Article Graph}

As discussed in Introduction, it is important to aggregate scattered travel descriptions of the celebrity from various news articles. We draw inspirations from word graphs designed by \citet{deng2019learning} and extend them to include articles as a Word-Article graph $\boldsymbol{\mathcal{G}_{loc}}$ ((a) in Figure~\ref{fig:framework}), which allows us to capture the semantic relations across multiple articles. The Word-Article graph is constructed as follows.

Given the celebrity $c$, date $d$, and the candidate location $loc$, we first obtain a subset of news articles $\mathcal{A}_{c,d,loc}$ from $\mathcal{A}_{c,d}$ by selecting articles that contain location $loc$. 
We take both articles and words in $\mathcal{A}_{c,d,loc}$ as nodes in the Word-Article graph $\boldsymbol{\mathcal{G}_{loc}}$, and initialize them using TF-IDF vectors and Word2Vec, respectively. To capture dependencies between words that are far apart, we use a sliding window to construct edges between words within sentences~\cite{deng2019learning}. To aggregate semantic information across different articles, we add edges between articles and words based on their co-occurrence within the articles~\cite{wang2020heterogeneous}. Meanwhile, we also record the word node subscripts of words that appear in the names of the location $loc$ and celebrity $c$ during Word-Article graph construction as the location word node $lw$ and celebrity word node $cw$, which will be used later in \textbf{Oriented Pooling}.

To identify important words from lengthy text descriptions, we use the graph attention network (GAT)~\cite{velikovi2017graph}, which employs an attention mechanism during graph aggregation, to selectively gather information from nearby nodes.

Given the Word-Article graph $\boldsymbol{\mathcal{G}_{loc}}$, its adjacency matrix $\mathbf{B}_{loc}\in \mathbb{R}^{(\iota+\tau)\times (\iota+\tau)}$, its word feature matrix $\mathbf{X}^{W}_{loc}\in \mathbb{R}^{\iota\times F_{W}}$ and the article feature matrix $\mathbf{X}^{A}_{loc}\in \mathbb{R}^{\tau\times F_{A}}$, where $\iota$ is the number of words and $\tau$ is the number of articles, we project the article and word nodes onto the same feature dimension using two independent linear layers with identical output dimension (Eq.~\eqref{eq1}-~\eqref{eq2}).
\begin{align}
    \mathbf{H}_{loc}^{W} &= \mathbf{W}_{W}\mathbf{X}_{loc}^{W}+b_{W},\label{eq1}\\
    \mathbf{H}_{loc}^{A} &= \mathbf{W}_{A}\mathbf{X}_{loc}^{A}+b_{A},\label{eq2}
\end{align}
where $\mathbf{W}_{*}$ is a learnable weight matrix. Hence, we get the node feature matrix $\ddot{\mathbf{H}}_{loc}=\mathbf{H}^{W}_{loc}\oplus \mathbf{H}^{A}_{loc}$ by concatenating the word feature matrix and the article feature matrix.

Then we use GAT to update the node representation. Since the module would be repeatedly utilized in the model, we omit the superscripts indicating layer number for brevity.


\begin{align}
    \mathbf{H}_{loc} = \textrm{GAT}(\ddot{\mathbf{H}}_{loc}, \boldsymbol{\mathcal{G}_{loc}}).
\end{align}

\subsubsection{Oriented Pooling} After updating node representations in the Word-Article graph, we design Oriented Pooling to focus on the semantics of the target celebrity and candidate location, thus reducing the distraction from irrelevant information. Oriented Pooling refines the Word-Article graph through the relevance of each node to the location word node $lw$ and the celebrity word node $cw$.

In Oriented Pooling, we first refer to SAGPool~\cite{lee2019self} to calculate the self-attention scores $s_{attn}$ on each node using graph convolution. To capture the semantic relation between the target celebrity and candidate location, we compute the similarity $s_{sim}$ between each node and the target nodes (location word $lw$ and celebrity word $cw$). Here we obtain the embeddings of target nodes by node subscripts recorded during Word-Article graph construction.
After combining $s_{attn}$ and $s_{sim}$ with additive attention, we get the oriented score $s_{ori}$.
\begin{align}
    s_{loc} &= \textrm{similarity}(\mathbf{H}_{loc},\mathbf{h}_{lw}),\label{eq:5sim}\\
    s_{c} &= \textrm{similarity}(\mathbf{H}_{loc},\mathbf{h}_{cw}),\label{eq:6sim}\\
    s_{sim} &= s_{loc} \odot s_{c}, \\ 
    s_{ori} &= \textrm{tanh}(\mathbf{W}_{\alpha}[s_{attn} \oplus s_{sim}]),
\end{align}
where $\mathbf{h}_{lw}$ and $\mathbf{h}_{cw}$ are the embeddings of node $lw$ and node $cw$, similarity(*) is the function to compute node similarity, $\odot$ is the element-wise product, and $\oplus$ means ``concatenate''.

Then we refine the graph using the oriented score. The pooling ratio $\epsilon\in (0,1)$ is a hyperparameter that determines the ratio of nodes to be saved. We select $\lceil \epsilon N\rceil$ nodes with the highest oriented score. $N$ is the number of nodes in the Word-Article graph.
\begin{align}
    idx &= \textrm{top-rank}(s_{ori},\lceil \epsilon N \rceil),\\
    \mathbf{H}_{loc} &= \mathbf{H}_{loc}[idx];\mathbf{B}_{loc}= \mathbf{B}_{loc}[idx],\\
    \mathbf{h}^{\diamond}_{loc} &= \textrm{Maxpooling}(\mathbf{B}_{loc},\mathbf{H}_{loc}),
\end{align}
where $\textrm{top-rank}(*)$ is the function to return the subscripts of the selected nodes and $\mathbf{B}$ is the adjacency matrix.
We operate GAT and Oriented Pooling iteratively in the location embedding learning module, and finally obtain the location embedding $\mathbf{h}_{loc}^{\diamond}$ of the candidate location $loc$ by max pooling.


After performing the location embedding learning module for each candidate location, we get the location node feature matrix $\textbf{H}^{\diamond}_{loc} = \left[ \mathbf{h}_{loc_1}^{\diamond}, \cdots, \mathbf{h}_{loc_k}^{\diamond} \right]^T \in \mathbb{R}^{k \times F}$, where $k = |Loc_{c,d}|$ is the number of candidate locations on date $d$ for the celebrity $c$, and $F$ is the feature dimension.

\subsection{Entity Embedding Learning}

As previously discussed, implicit trips pose another challenge in celebrity trip detection. To utilize the background knowledge of celebrities and other named entities, we first employ SpaCy\footnote{\url{https://spacy.io/}} to recognize entities of specific types, including person, nationalities or religious or political groups (NORP), facility and organization, from $\mathcal{A}_{c,d}$. Then we obtain the structured information and relationships of entities from Wikidata embeddings provided by OpenKE~\cite{han2018openke} through a strict text matching method such that we do not introduce noise in this process.


Given the entity $ent$ and its embedding $h_{ent}$ initialized by Word2Vec, we construct an Entity Sub-graph $\boldsymbol{\mathcal{G}_{ent}}$ specific to $ent$ based on the relation data from OpenKE ((b) in Figure~\ref{fig:framework}). Nodes in $\boldsymbol{\mathcal{G}_{ent}}$ are $ent$ and external entities connected to $ent$ through the data provided by OpenKE. These relations are represented as edges in the graph. To learn the entity embedding vector, we adopt the CompGCN~\cite{vashishth2019composition} layer, which is the state-of-the-art model for learning entity and relation embeddings on graphs.

\begin{align}
    \bar{\mathbf{h}_{ent}} &= \sigma(\sum_{(r,u)\exists (ent,r,u)\in \boldsymbol{\mathcal{G}_{ent}}}\mathbf{W}_{es}\phi(\mathbf{h}_{u},\ell_{r})),\label{compgcn} \\
    \ell_{r} &= \mathbf{W}_{edge}\cdot \ell_{r}.
\end{align}
where $\mathbf{h}_{u} $ and $\ell_{r}$ denote features for node $u$ and relation type $r$ in the $\boldsymbol{\mathcal{G}_{ent}}$, respectively. $\mathbf{W}_{es}$ is the weight parameter for aggregating node and edge features, and $\phi(\cdot)$ is the multiplication operator. $\sigma(\cdot)$ is the activation function. $\mathbf{W}_{edge}$ is the weight parameter for updating the relation type embedding.

After aggregating the features of nodes and edges, we obtain the semantic feature $\bar{\mathbf{h}_{ent}}$ from the sub-graph. Then we concatenate the initialized embedding of entity $\mathbf{h}_{ent}$ with the semantic feature $\bar{\mathbf{h}_{ent}}$, and obtain the updated embedding of entity $ent$ through a non-linear transformation:
\begin{align}
    \mathbf{h}_{ent}^{\clubsuit} = \sigma(\mathbf{h}_{ent} \oplus \bar{\mathbf{h}_{ent}}).
\end{align}
The entity embeddings $\mathbf{H}^{\clubsuit}_{ent} = \left[\mathbf{h}_{ent_1}^{\clubsuit}, \cdots, \mathbf{h}_{ent_{n}}^{\clubsuit} \right]^T \in \mathbb{R}^{n\times F}$ are obtained. $n$ is the number of entities.

\subsection{Event Embedding Learning}
Events also provide important information for detecting implicit trips, since celebrity trips are often associated with specific events~\cite{doherty2009potus,ostrander2019presidents} and our analysis on Trip Dataset gives consistent evidences (see Distinguishing Ability of Events).
However, pre-trained embeddings from external knowledge bases cover limited events. To capture the current states of events, we propose a module that integrates contemporaneous event-related news articles to obtain the event embeddings.


Formally, given the event entity $eve$ recognized by SpaCy, the framework retrieves sentences $\mathcal{S}_{eve}$ containing the event $eve$ from news articles published on the date $d$. We introduce an attention mechanism that differentiates the importance of sentences by assigning varying weights to their content and aggregates the feature of sentences based on weights, since not all sentences contribute equally to the status of events.

Specifically, we encode the $i$-th sentence in $\mathcal{S}_{eve}$ and get the sentence vector $\mathbf{v}_i$ by Word2Vec. Afterwards, we derive the attention values by feeding the sentence vector $\mathbf{v}_{i}$ into an attention layer, followed by obtaining the normalized attention weight $\lambda_{i}$ through a softmax function. Finally, the representation $\tilde{\mathbf{h}_{eve}}$ is obtained as a weighted sum of each sentence vector, and the attention layer can be trained end-to-end while gradually learning to assign more attention to reliable and informative sentences.
\begin{align}
    \mu_{i} &= \textrm{Sigmoid}(\mathbf{W}_{eve}\mathbf{v}_{i} + b_{eve}),\\
    \lambda_{i} &=\frac{\textrm{exp}(\zeta_{i}\mu_{i})}{\sum_{j}\textrm{exp}(\zeta_{j}\mu_{j})},\\
    \tilde{\mathbf{h}_{eve}} &=  \sum_{i}\lambda_{i}\mathbf{v}_{i},
\end{align}

where $\zeta_{i}$ is the parameter for $\mathbf{v}_{i}$ and $\mathbf{W}_{eve}$ is the parameter matrix. Intuitively, the number of news articles related to an event increases when the event is ongoing, thus reflecting the state of the event. Therefore, we add a vector representing daily counts of news articles that contain the event.
When the framework detects event $eve$ on date $d$, it counts the number of articles containing event $eve$ from $d-\mathcal{Q}$ to $d+\mathcal{Q}$ to construct the vector $\mathbf{z} \in \mathbb{R}^{1\times (2\mathcal{Q}+1)}$, where $\mathcal{Q}$ is the date offset. Then, the framework applies a linear layer to project the vector to a low-dimensional space $\mathbf{h}_{\mathbf{z}}=\psi_{1}(\mathbf{z})$ and fuses it with $\tilde{\mathbf{h}_{eve}}$ to get the event status embedding $\mathbf{h}_{eve}^{\spadesuit}$.
\begin{align}
    \mathbf{h}_{eve}^{\spadesuit} = \sigma(\psi_{2}(\mathbf{h}_{\mathbf{z}}\oplus \tilde{\mathbf{h}_{eve}}))+\mathbf{h}_{\mathbf{z}}\label{eventstatus},
\end{align}
where $\sigma(\cdot)$ is activation function and $\psi_{*}(\cdot)$ is linear layer. In addition, we design a skip connection of the article-amount embedding $\mathbf{h}_{\mathbf{z}}$ to allow its gradients to flow in the network. Based on the above operation, we get the event embeddings $\mathbf{H}^{\spadesuit}_{eve} = \left[ \mathbf{h}_{eve_1}^{\spadesuit}, \cdots, \mathbf{h}_{eve_m}^{\spadesuit} \right]^T \in \mathbb{R}^{m\times F}$ ($m$ is the number of events) for all events.

\subsection{Trip Graph Learning}

To perform trip detection from a comprehensive perspective and jointly train all modules, the framework consolidates the embeddings of locations, entities, and events from the same date in one graph. With the final representations learned from this module, locations can then be classified.

\subsubsection{Trip graph construction} 


Formally, we define the Trip Graph of the celebrity $c$ on date $d$ as $\boldsymbol{G_{c,d}} = (\mathcal{V},\mathcal{E})$ ((d) in Figure~\ref{fig:framework}), where $\mathcal{V}$ consists of candidate locations $Loc_{c,d}=\left\{loc_{1},...,loc_{k}\right\}$, entities $Ent=\left\{ent_{1},...,ent_{n}\right\}$ and events $Eve =\left\{eve_{1},...,eve_{m}\right\}$. We build an edge between the location node $loc_{k}$ and the entity node $ent_{n}$ (the event node $eve_{m}$), if entity $ent_{n}$ (events $eve_{m}$) appears in $\mathcal{A}_{c,d,loc_{k}}$.

\subsubsection{Trip graph aggregation}
Given the location embeddings $\mathbf{H}^{\diamond}_{loc}$, entity embeddings $\mathbf{H}^{\clubsuit}_{ent}$, event embeddings $\mathbf{H}^{\spadesuit}_{eve}$, and the Trip Graph $\boldsymbol{G_{c,d}}$, the framework jointly trains the embeddings of locations, entities, and events in the Trip Graph aggregation module. The Trip Graph aggregation module propagates the gradient back to the location embedding learning module, the entity embedding learning module, and the event embedding learning module.

Specifically, we concatenate embeddings of locations, entities, and events, and obtain the feature matrix $\hat{\mathbf{H}}_{c,d} \in \mathbb{R}^{(k+n+m)\times F}$ of the Trip Graph $\boldsymbol{G_{c,d}}$.
\begin{align}
    \hat{\mathbf{H}}_{c,d} = \mathbf{H}^{\diamond}_{loc}\oplus \mathbf{H}^{\clubsuit}_{ent} \oplus \mathbf{H}^{\spadesuit}_{eve}.
\end{align}
To capture the various influences of entities and events on trip detection, the framework uses GAT to aggregate the information of the Trip Graph.
\begin{align}
    \mathbf{H^{\heartsuit}_{c,d}} = \textrm{GAT}(\mathbf{\hat{H}}_{c,d},\boldsymbol{G_{c,d}}).
\end{align}


Afterwards, our framework derives the feature matrix $\textrm{H}_{c,d}^{\heartsuit}$ of the Trip Graph $\boldsymbol{G_{c,d}}$ and applies the softmax function to the location nodes' features to estimate the probability of each location being the celebrity trip location.
\begin{align}
    y_{i}^{'} = \textrm{Softmax}(\mathbf{h}_{loc_{i}}^{\heartsuit} ).
\end{align}

The framework compares the prediction result with the ground truth and optimizes via the binary cross-entropy loss:

\begin{align}
    \mathcal{L} =-\frac{1}{|Loc_{c,d}|} \sum_{i}^{|Loc_{c,d}|}(y_{i}  \log(y^{'}_{i})+(1-y_{i}) \log(1-y^{'}_{i})).
\end{align}
where $|Loc_{c,d}|$ is the number of candidate locations on date $d$, and $y_{i}$ is the ground truth label of $i$-th location.



\subsection{Extraction Tools for Dates and Locations}

As mentioned in Problem Statement, we need to extract dates and locations from text. Existing time extraction tools, such as SUTIME~\cite{chang2012sutime}, are general-purposed for various time information, while we only care about dates. As a result, these tools are quite slow and may not archive the best performance (see Evaluation of Extraction Tools). We therefore develop a rule-based \textbf{time extraction tool} with regular expressions to handle absolute dates (e.g., ``July 16, 2022'') and relative dates (e.g., ``yesterday'').

Regarding locations, we can potentially use the results from NER tools. Again, as general-purpose tools, they often produce coarse results (e.g., ``North Korea's'' instead of ``North Korea'') and locations with containment relationships (e.g., ``New York'' v.s. ``United States'') that affect further modeling work. Thus, we develop a \textbf{location extraction tool} that integrates GeoNames\footnote{\url{https://www.geonames.org/}}, a geographic database, to get high-quality location names, and uses administrative-level information from OpenStreetMap\footnote{\url{https://www.openstreetmap.org/}} to resolve containment relationship issue.

\section{Experiments}

\subsubsection{Data Collection}
We construct a ground truth trip dataset from the Wikipedia pages that list celebrity trips\footnote{\url{https://en.wikipedia.org/wiki/List_of_international_prime_ministerial_trips_made_by_Theresa_May} The Wiki page for Theresa May's trips, as an example.}. In total, there are 5,687 trips with dates and locations made by 26 politicians and 24 artists, from Jan 2016 to Feb 2021. 

We then collect the URLs of news articles published during this period of time from the GDELT project\footnote{\url{https://www.gdeltproject.org/}} and use Newspaper3k\footnote{\url{https://github.com/codelucas/newspaper}} to acquire the news content of the URLs.

\subsubsection{Data Processing and Labeling}
To obtain the dataset for training and testing in the format described in the Problem Statement, we further process the collected data. As a pre-step, we first perform NER on the news articles using SpaCy.
For each ground truth trip, we extract candidate locations of the corresponding celebrity from news articles that contain the celebrity name and relate to the trip date (published on the date, or mentioning the date in text) by our extraction tool. For each candidate location, if it is actually visited on that date according to Wikipedia, it will be labelled positive and vice versa. For each negative/positive location, its articles are news articles that contain the location name among the above news articles. As a result, we identify 2,404 locations as ``positive locations'' and 9,707 locations as ``negative locations''. This approach allows us to construct the \textbf{Trip Dataset} for the task, with each row recording the celebrity name, location name, date, article IDs, and the label. Table~\ref{tab: dataset example} shows a sample from this dataset. 


\begin{table}[htb]
\setlength\tabcolsep{3pt}
  \centering
    \begin{tabular}{ccccc}
    \toprule
  Celebrity& Location &Date&Articles&Label\\
    \midrule
  Donald Trump &Langley &01/21/2017 &$\mathcal{A}_{1}$,$\mathcal{A}_{2}$,...&Positive\\
\bottomrule
    \end{tabular}%
    \caption{A sample from the Trip Dataset.}
    \label{tab: dataset example}
\end{table}

\subsubsection{Train/Test Split}
With July 1, 2019 as the split date, we partition the Trip Dataset into training (66.5\%) and testing (33.5\%) to avoid potential data leakage, since SpaCy is trained on data prior to July 2019. Then we randomly sample 10\% from the training set as the validation set to optimize the hyperparameters. Table~\ref{tab: dataset} lists the details of the training and test sets.


\begin{table}[htbp]
\setlength\tabcolsep{3pt}
  \centering
    \begin{tabular}{cccc}
    \toprule
   Trip Dataset& Positive & Negative &Period\\
    \midrule
   Train &1,715 &6,341&01/01/2016 - 06/30/2019\\
   Test &689 &3,366&07/01/2019 - 02/28/2021\\
\bottomrule
    \end{tabular}%
    \caption{Training and test sets from the Trip Dataset.}
    \label{tab: dataset}%
\end{table}%

\subsubsection{Distinguishing Ability of Events}

We wonder whether the positive locations are associated with different events, compared with negative ones, as suggested by some studies~\cite{doherty2009potus,ostrander2019presidents}. If so, event information may be helpful for distinguishing positive locations from negative ones. In Table~\ref{tab:eve_freq}, we list the top 10 most frequent events extracted by SpaCy from articles of positive samples and negative samples respectively, and see how they differ. We can see that 6 events out of 10 for positive samples are international meetings of leaders (highlighted in Table~\ref{tab:eve_freq}), while for the negative ones, the events seem a mixture of protests (``Women's March''), sports events (e.g., ``Olympics''), social gatherings (``Unite the Right''), and disasters (e.g., ``Hurricane Harvey'' and ``Sandy Hook''). This hints that event information might contain clues for classifying the samples.


\begin{table}[htbp]
\setlength\tabcolsep{3pt}
  \centering
    \begin{tabular}{cc}
    \toprule
    Positive & Negative \\
    \midrule
    \textbf{G20} & Hurricane Harvey\\
    \textbf{World Economic Forum} & Women's March\\
    \textbf{G7} & Unite the Right\\
    Brexit & Republican National Convention\\
    \textbf{NATO Summit} & Democratic National Convention\\
    Hurricane Harvey & Olympics\\
    \textbf{APEC Summit} & Super Tuesday\\
    Women's March & Super Bowl\\
    World Cup & Hurricane Maria\\
    \textbf{Belt and Road Forum} & Sandy Hook\\
\bottomrule
    \end{tabular}%
    \caption{Frequent events associated with positive and negative samples.}
    \label{tab:eve_freq}
\end{table}%

\subsubsection{Implementation Details}
\label{sec:implement}
To prepare the news articles for further processing, we remove stop words and perform stemming by NLTK\footnote{\url{https://www.nltk.org/}}. In the location embedding learning module and event embedding learning module, to get original representations of words and sentences, we train an unsupervised CBOW Word2Vec~\cite{mikolov2013efficient} with the dimension of 100 on the entire news dataset.


We describe the parameters for the four components:
\begin{enumerate}[label=\alph*)]
    \item Location Embedding Learning:
    During the construction of Word-Article graphs, we set the size of the sliding window to 15 and aggregate the graph with two GAT layers. Additionally, the dimension of TF-IDF vectors used to initialize article features is 1,000. In Oriented Pooling, we use cosine similarity as the similarity function in Eq.~\eqref{eq:5sim} and Eq.~\eqref{eq:6sim}.
    \item Entity Embedding Learning:
    We utilize pre-trained Wikidata embeddings provided by OpenKE~\cite{han2018openke} to construct the Entity Sub-graphs, with entities and relations dimensions set to 50. One CompGCN layer is employed in Entity Sub-graphs.
    \item Event Embedding Learning:
    The date offset $\mathcal{Q}$ is set to 7.
    \item Trip Graph Learning:
    We set the input dimension $F$ of Trip Graph to 128 and the number of GAT layers to 2.
\end{enumerate}
To train CeleTrip, we set the learning rate to 0.001 and use the Adaptive Moment Estimation (Adam) optimizer~\cite{kingma2014adam}. Meanwhile, we adopt an early stop strategy to avoid overfitting.


\subsubsection{Evaluation Metrics}
\label{sec:evaluation metrics}
We use Accuracy (Acc), Precision (P), Recall (R), and F1 score for evaluation.
Precision measures the proportion of the locations predicted as positive that are actually visited by the target celebrities, while Recall indicates the ratio of the actually visited locations that are correctly classified. F1 score, as the harmonious average of Precision and Recall, is an important overall metric in our scenario.

\subsubsection{Baselines} 

Here we introduce seven baselines. The first two (LocFre and LocJaccard) are based on the intuition described in Introduction which relies on the frequency of each candidate location to make the decisions. The next four models, LR (TF-IDF), CNN and Bi-LSTM, and GCN, as mentioned in Introduction, are often used in PRED and other Detection tasks. Among them, the graph-based GCN may bring improvements, since it has the ability to capture long-distance dependencies. But as discussed before, GCN employs max pooling to read out the graph, compared with our model CeleTrip. We need to mention that Oriented Pooling in our model uses the features of celebrity/location word nodes during graph pooling, and this information is not originally included in the input to these four models. Therefore, as a fair comparison, we concatenate the celebrity name and location name embeddings with the output vectors of these four models, before feeding them to the classifying modules. 

Another important aspect that we want to evaluate is the inclusion of external knowledge through entity embeddings and event embeddings. The aforementioned four models by their original designs do not incorporate such information, and therefore we alternatively replace the location embedding learning module in our framework with each of these models, to see how the knowledge and the joint learning scheme would affect their performance. For the same purpose, we introduce CeleTrip$_\textrm{\ w/o\ kn}$, without modules for external knowledge and the accompanying joint learning scheme, to compare against the full CeleTrip.


\begin{itemize}
    \item \textbf{LocFre:} It chooses the location mentioned most frequently in the news as the trip location.
    \item \textbf{LocJaccard:} LocFre favors the locations that are generally popular across all news articles and this can be a bias. To normalize the frequency by popularity, LocJaccard calculates the Jaccard coefficient between articles containing candidate location and articles containing target celebrity, as a possible improvement over LocFre.
    \item \textbf{LR (TFIDF)~\cite{dickinson2015identifying}:} 
    The researchers use TF-IDF vectors as the representations of news articles and employs Logistic Regression as the classifier for a PRED task. We directly adopt it in our experiments.
    \item \textbf{CNN~\cite{nguyen2017robust}:} 
    In their study of detecting crisis from personal tweets, the model learns local semantic features of articles. The article feature vectors yielded by the CNN model are then concatenated together as the input to the classifier. In our experiments, we use the softmax classifier for all deep learning models.
    \item \textbf{Bi-LSTM~\cite{yen2019personal}:} 
    For the PRED task, the researchers use Bi-LSTM, a sequential learning model to captures the continuous semantic features of Chinese characters in tweets. In our task, we treat English words as the Chinese characters in their study.
    \item \textbf{GCN~\cite{deng2019learning}:} In the event prediction task, the researchers model articles as word graphs and aggregate the information in the word graph by graph convolutional network (GCN). In the experiments, we construct the same Word-Article graphs instead of the original word graphs.
    \item \textbf{CeleTrip$_\textrm{\ w/o\ kn}$:} This model employs the proposed Word-Article graph to model location context and obtains the representation of the graph via Oriented Pooling.
\end{itemize}

\subsection{Experimental Results}
\label{Experimental Results}
\paragraph{Prediction Performance}
The results averaged over 5 runs are presented in Table~\ref{tab: Method result}.
Among the models, CeleTrip exhibits the best overall performance (F1=$82.53\%$), followed by CeleTrip$_\textrm{\ w/o\ kn}$, the version of our model without external knowledge (entity and event embeddings).

\begin{table}[!htb]
  \centering
    \begin{tabular}{lllll}\toprule
    Methods&F1 (\%)&P (\%)&R (\%)&Acc (\%)\\\midrule
 LocFre& 37.34 & 39.06&  35.76  &81.96\\
LocJaccard&  40.65&  40.59 &40.53&85.37\\
    \midrule
    LR (TFIDF) &	64.65&	\textbf{88.64} &	52.83 &	90.69\\
    CNN& 63.75	&	66.46 &61.25&89.72\\
    Bi-LSTM&66.93 &	71.95& 62.55 &	90.26\\
    GCN	& 60.18	&60.71& 59.65&86.58\\
    CeleTrip$_\textrm{\ w/o\ kn}$ &\textbf{74.11}& 78.89 &\textbf{69.88}&\textbf{92.04}\\

    \midrule
     \multicolumn{5}{l}{With External Knowledge}\\
   LR (TFIDF) & 68.50&84.14& 57.76&	91.29\\ 
    CNN	& 69.85 &	79.08 &62.55&90.83 \\ 
    Bi-LSTM&70.68 &	81.71 &62.26 &	91.22\\ 
    GCN & 64.70&	71.94 &58.78&	89.10\\ 

 \textbf{CeleTrip}&\textbf{82.53}&\textbf{86.17} &	 \textbf{79.27} &\textbf{94.30} \\
\bottomrule
    \end{tabular}%
    \caption{Performance comparison on the test set.}
    \label{tab: Method result}%
\end{table}%

As the best model among these without external knowledge (upper part of the table), CeleTrip$_\textrm{\ w/o\ kn}$ outperforms by large margins, suggesting the effectiveness of graph structure with attention mechanism. The GCN model which is also graph-based, however, does not perform so well, only surpassing the two frequency-based methods. A possible explanation is that words should not be treated equally in our scenario -- indicative words need to have higher weights. This is further supported by the interesting observation that the LR (TFIDF) method, has the highest Precision (88.64\%) and it gives higher weights to indicative words relevant to trips, such as ``visit'' or ``arrive''. On the other hand, LR (TF-IDF) has the lowest Recall, and it may be related to the fact that it does not model relationships between words, therefore missing meaningful word dependencies/combinations.


We notice that CeleTrip$_\textrm{\ w/o\ kn}$ outperforms CNN and Bi-LSTM, which focus on local semantics, by 10.36\% and 7.18\% in F1, highlighting the necessity of capturing long-distance dependencies in our scenario. In our task, descriptions of a trip sometimes scatter in different sentences or articles, making it important to consider word relationships that are far apart.


Additionally, we observe that simple frequency-based models have F1 scores of only 40\%, indicating limited practicality despite some predictive effect. Conversely, models incorporating textual information significantly surpass naive approaches in all metrics, demonstrating the importance of textual description in trip detection.

\paragraph{Performance for Unseen Celebrities in Training}
Two politicians, ``Boris Johnson'' and ``Joe Biden'', appear only in the test set, with 33 possible trips (16 positives and 17 negatives). All of these trips are correctly classified by CeleTrip, indicating the model generalizes well for these two celebrities.

\paragraph{Effectiveness of External Knowledge}

The performance of all baseline models improves when combined with external knowledge. Specifically, the incorporation of external knowledge in LR (TFIDF), CNN, Bi-LSTM, GCN, and CeleTrip$_\textrm{\ w/o\ kn}$ yields improvements of 3.85\%, 6.10\%, 3.75\%, 4.52\%, and 8.42\% respectively in F1 score, among which CeleTrip has the largest gain.

\subsection{Ablation Study}

We remove certain component(s) in CeleTrip to validate their efficacy, with the results in Table~\ref{tab: Ablation result}. CeleTrip$_\textrm{\ w/o\ ent}$ removes the entity embedding learning module and directly uses pre-trained entity (Word2Vec) embeddings. Similarly, CeleTrip$_\textrm{\ w/o\ eve}$ gets rid of the event embedding learning module and uses pre-trained event embeddings constructed from Word2Vec embeddings  (see construction details in Method). CeleTrip$_\textrm{\ w/o\ ent\&eve}$ is the ablation without both aforementioned modules and employs original entity and event embeddings as CeleTrip$_\textrm{\ w/o\ ent}$ and CeleTrip$_\textrm{\ w/o\ eve}$ do. Additionally, we replace Oriented Pooling in CeleTrip with max pooling  and obtain CeleTrip$_\textrm{\ w/o\ ori}$.



In Table~\ref{tab: Ablation result}, as expected, CeleTrip has the best overall performance compared with all variants, and CeleTrip$_\textrm{\ w/o\ ent\&eve}$ is the worst, demonstrating the effectiveness of corresponding module designs. Compared with CeleTrip, CeleTrip$_\textrm{\ w/o\ eve}$ has a slightly higher Precision (increased by 1.05\%) but a much lower Recall (decreased by 5.98\%). It is possible that by introducing event information from additional news articles, the Recall is greatly improved, but the accompanying noise affects Precision by a small margin. Moreover, we find that CeleTrip outperforms CeleTrip$_\textrm{\ w/o\ eve}$ by 2.88\% and CeleTrip$_\textrm{\ w/o\ ent}$ by 4.75\% in the F1 metric, indicating that the entity embedding learning module has a larger contribution, compared with the other module. It is worth noting that Oriented Pooling brings a large improvement of 5.73\% in F1, and the enhancement may be attributed to the suppression of unrelated information.

\begin{table}[!htb]
  \centering
    \begin{tabular}{lllll}\toprule
     Methods&F1 (\%) &P (\%)&R (\%)&Acc (\%)\\\midrule
    CeleTrip$_\textrm{\ w/o\ ent\&eve}$& 76.68 &	84.20& 70.39&92.77\\
    CeleTrip$_\textrm{\ w/o\ ori}$& 76.80 &	81.29& 72.91 &92.51\\
    CeleTrip$_\textrm{\ w/o\  ent}$&77.78& 84.38& 72.13&93.00\\
    CeleTrip$_\textrm{\ w/o\ eve}$ &79.65&	\textbf{87.22} &73.29&93.64\\
   \textbf{CeleTrip}&\textbf{82.53}&86.17 &	 \textbf{79.27} &\textbf{94.30} \\
\bottomrule
    \end{tabular}%
    \caption{Performance of different CeleTrip variants.}
    \label{tab: Ablation result}%
\end{table}%

\subsection{Parameter Sensitivity}


\subsubsection{Window Size for Word-Article Graph} When building the Word-Article graph, we have a sliding window to decide which words should be connected. Figure~\ref{window size} shows that 15 may be an ideal size -- smaller windows may fail to capture long-distant dependencies while the ones larger than 15 may introduce more noise than the possible gains.


\subsubsection{Node Dimension $F$ in Trip Graph} 
In Figure~\ref{dim F}, the model performance improves with increased input dimensionality. Considering the fact that the improvement is marginal beyond $F=128$, and the trade-off between performance and computational cost, we decide $128$ to be a practical cut.



\subsubsection{Layers of Graph Aggregators} 
\begin{figure}[!htbp]
\centering
\subfigure[Window Size]{
\begin{minipage}[!htb]{0.5\linewidth}
\centering 
\includegraphics[width=1\textwidth]{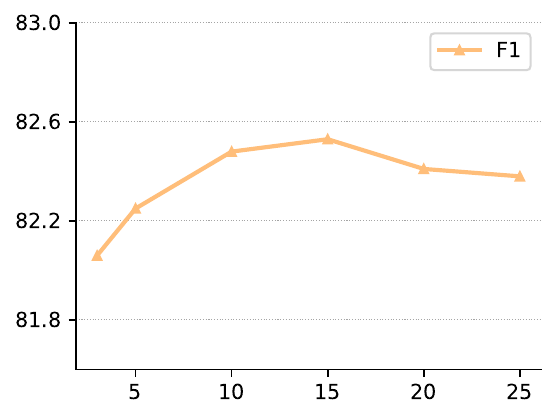}
\label{window size}
\end{minipage}%
}%
\subfigure[Dimension $F$]{
\begin{minipage}[!htb]{0.5\linewidth}
\centering 
\includegraphics[width=1\textwidth]{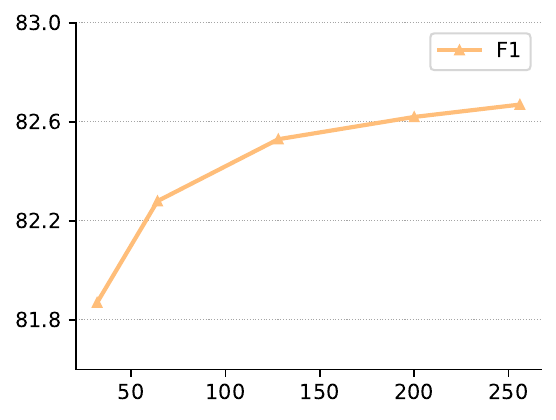}
\label{dim F}
\end{minipage}%
}%
\quad
\centering
\subfigure[\# of GATs in Word-Article graph]{
\begin{minipage}[!htb]{0.462\linewidth}
\centering 
\includegraphics[width=1\textwidth]{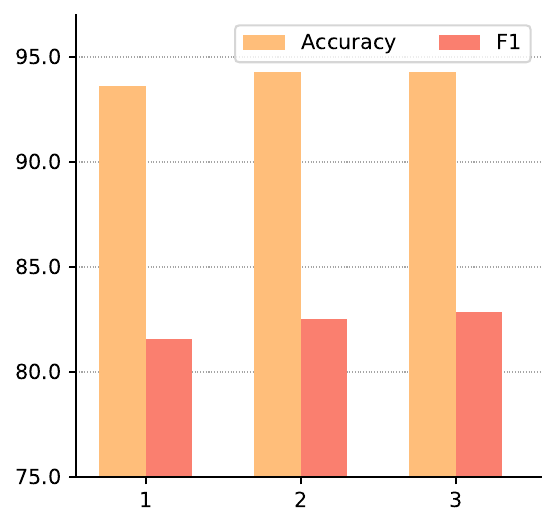}
\label{wa graph}
\end{minipage}%
}%
\subfigure[\# of GATs for Trip Graph]{
\begin{minipage}[!htb]{0.462\linewidth}
\centering 
\includegraphics[width=1\textwidth]{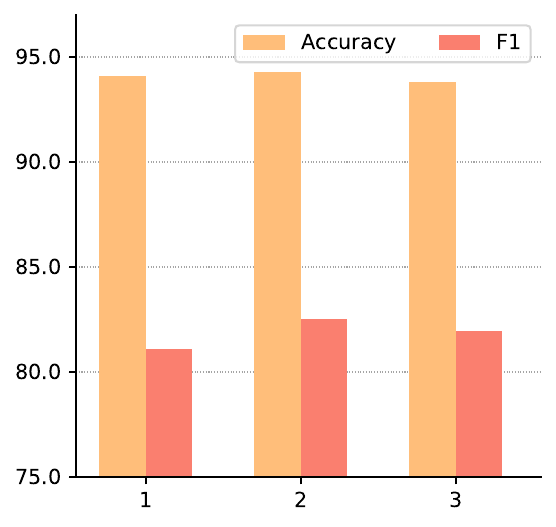}
\label{trip graph}
\end{minipage}%
}%
\centering
\caption{Sensitivity analysis of different parameters. The horizontal coordinates indicate the values of parameters, and the vertical coordinates indicate the value of corresponding metrics.}
\label{fig: Hyperparameter}
\end{figure}

We vary the number of GAT layers from 1 to 3 for the Word-Article graph and the Trip Graph, and show the results in Figure~\ref{wa graph} and Figure~\ref{trip graph}, respectively. It seems that 3 is a good choice for the Word-Article graph, while 2 may be ideal for the Trip Graph. Observing the increasing performance in the Word-Article graph, as the number of layers increases, we guess that for a large graph like the Word-Article graph, more layers may help encode more helpful information, and for a much smaller Trip Graph, too many layers may cause over-smoothing~\cite{chen2020measuring}.


\subsection{Qualitative Evaluation}
We examine what the event and entity embedding learning modules capture respectively. Besides, we perform an error analysis on incorrectly classified samples.

\subsubsection{Interpretation of Event Embeddings}
\begin{figure}[!htb]
    \centering
    \includegraphics[width=\linewidth]{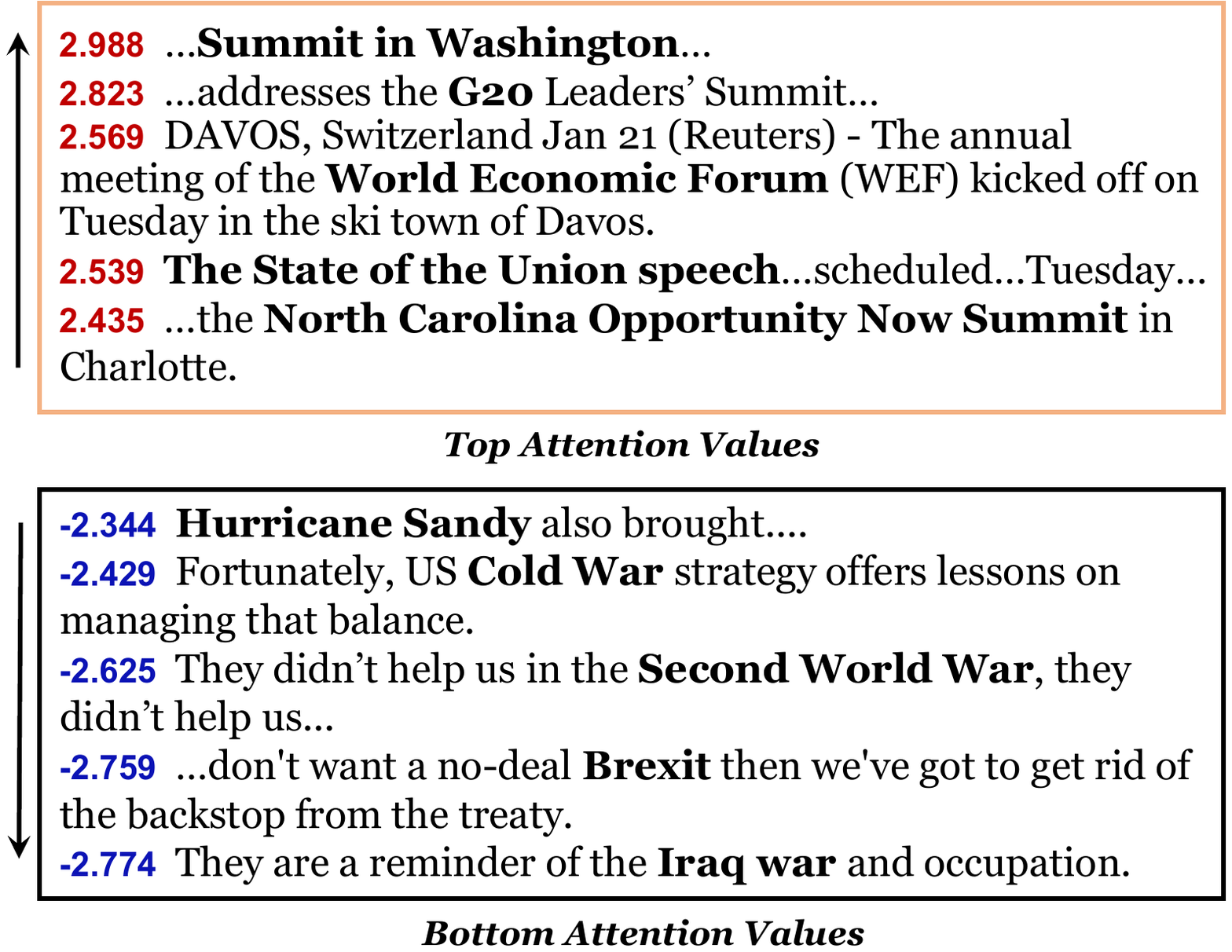}
    \caption{Top 5 and bottom 5 sentences ranked by attention values from the event embedding learning module.}
    \label{fig:eve_attn}
\end{figure}

The event embedding learning module learns the representations of the sentences that contain the events appearing in the context of candidate locations, during which each sentence receives an attention value. We manually examine top and bottom 500 sentences among 11,891 sentences ranked by their attention values, respectively. We find that in 85\% of the top sentences, the events are actually related to the positive trips. As a comparison, in 96\% of the bottom sentences, the corresponding events seem unrelated and are mostly about natural disasters and wars. As a demonstration, we show top 5 and bottom 5 sentences in Figure~\ref{fig:eve_attn}, and the events are highlighted.

\subsubsection{Celebrity Entity Embedding Visualization}
\begin{figure}[!htb]
    \centering
    \includegraphics[width=0.95\linewidth]{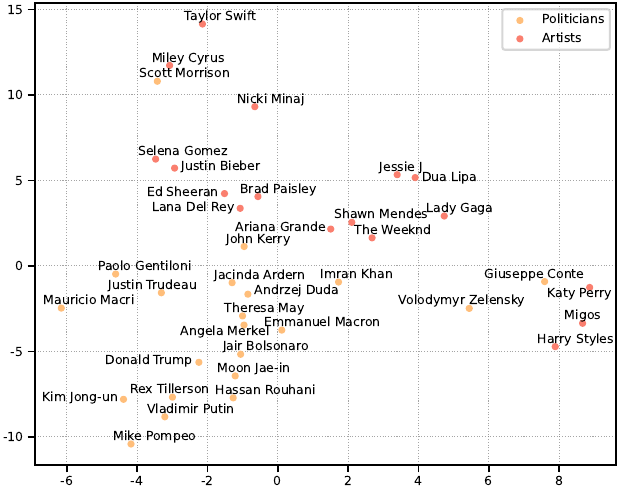}
    \caption{Visualization of the learned representations for celebrity entities, through t-SNE.}
    \label{fig:personenttiy}
\end{figure}

To see how the relationships between entities reflect the real world, we choose to study the celebrity entities by using t-SNE~\cite{van2008visualizing} to reduce the embedding dimensions and visualizing them in a 2-D space. In Figure~\ref{fig:personenttiy}, interestingly, politicians cluster together in the lower left diagonal half, while artists occupy the other half. The relatively close proximity between European politicians, compared with other regions in the world, may be attributed to frequent regional meetings within the EU. 
Specifically, Theresa May and Angela Merkel form the closest pair among all celebrities and this may be the result of frequent co-occurrences, given that the two had overlapped terms of office during which both focused on Brexit and other EU affairs. These highlight the model's potential to identify regional political patterns.

The artists in the upper right diagonal half, on the other hand, are more scattered than politicians. The reason for this discrepancy may be that artists are more likely to attend solo activities/performances, while politicians often meet each other. This is further confirmed by our statistics -- 50.15\% of the politician trips are joined by others, and the ratio is only 13.04\% for artists.


\subsubsection{Error Analysis}

We examine the error cases to summarize their causes. In the evaluation, there are two types of errors, false positives (locations not actually visited but predicted as visited, associated with Precision) and false negatives (locations actually visited but predicted as not visited, associated with Recall). We sample 25\% from each type, resulting in 21 false positives and 36 false negatives.

False positives are mainly caused by: (1) trips by other candidates (38\%); and (2) absent actual dates (19\%). Though Oriented Pooling is designed to put focus on target celebrities, trips of others still distract the model to some extent. On the other hand, actual dates of the trips may not appear in the articles, and other dates in the articles or the publishing dates may be mistakenly used instead.

The main error reasons for false negatives are: (1) implicit trips with very vague clues (43\%); and (2) the target celebrity and the visited place have no apparent associations, though appearing in the same article (37\%). For reason (1), we see cases such as campaigns and fund-raising events, indicating the target celebrities must have been somewhere, but the available information is not enough for the model to link to external knowledge. We demonstrate reason (2) by the following sentences from one article:
\textit{``Some leaders, including German Chancellor \textbf{Angela Merkel}, see regulating AI as part of the EU’s …''} and
\textit{``Von der Leyen said the \textbf{Brussels} summit deal, reached in the wee hours of Friday by 27 national EU leaders''}. The association between Merkel and Brussels seems very insufficient for the model to classify it as a positive trip.

\subsection{Evaluation of Extraction Tools}
\label{Sec:framework effectiveness}



\subsubsection{Time Extraction}


We compare our detection tool with SUTIME~\cite{chang2012sutime}, an advanced time recognition tool, on the TempEval dataset~\cite{uzzaman2012tempeval}.
Our tool achieves an accuracy of 91.1\%, which is slightly higher than SUTIME (90.6\%), and greatly reduces the processing time by over 80\%, from 41 mins to 8 mins.

\subsubsection{Location Extraction}
Our location extraction tool (SpaCy+GeoNames) is built upon the NER results from SpaCy, and further uses a geolocation database, GeoNames, to remove inaccurate location names and expand some abbreviated names to increase matches. The performance is evaluated on the New York and Christchurch datasets~\cite{middleton2013real}, and Table~\ref{tab: prelocation} shows the large improvements in F1 scores and Precision.


\begin{table}[!htbp]
  \centering
    \begin{tabular}{cccc}\toprule
    Methods & F1 (\%) & P (\%) & R (\%)\\
    \midrule
    \multicolumn{4}{c}{New York Dataset}\\
    \midrule
    SpaCy & 68.02 & 60.36 &  77.91\\
    SpaCy+GeoNames & \textbf{82.38} & \textbf{86.66} &  \textbf{78.50}\\
    \midrule
    \multicolumn{4}{c}{Christchurch Dataset}\\
    \midrule
    SpaCy & 64.88 & 63.50 &  \textbf{66.31}\\
    SpaCy+GeoNames & \textbf{71.18} & \textbf{78.62} & 65.02\\
\bottomrule
    \end{tabular}%
    \caption{Performance of location extraction strategies.}
    \label{tab: prelocation}%
\end{table}%





\section{Related Work}

We discuss three threads of research: celebrity trip analysis, person-related event extraction and graph neural network.

\subsection{Celebrity Trip Analysis}

Celebrity trips have purposes that may extend to important implications. For instance, politicians travel to negotiate foreign policies and appeal to the public~\cite{kernell2006going,ostrander2019presidents}, while artists make trips to promote their work~\cite{hautala2019creativity}. Based on the mobility (births and deaths) of 150,000 notable individuals, researchers carry out an impressive analysis to chart cultural history~\cite{schich2014network}.


Among the quantitative studies, U.S. presidential travels have been extensively studied. Findings include that the presidents' trips are influenced by world events and the level of U.S. involvement~\cite{cavari2019going}, and the purposes of early presidents' domestic trips are to advance re-election~\cite{sellers2006presidential}, promote policies and meet with key leaders or groups~\cite{doherty2009potus}.

As mentioned in Introduction, data scarcity is a common problem that prevents relevant studies from being extended to large-scale, fine-grained, and network-wise.


\subsection{Person-Related Event Detection}

In Introduction, we have briefly discussed Person-Related Event Detection (PRED), which classifies user-posted short text into predefined personal life events, while our classification task deals with much longer news articles. In a relatively early study~\cite{dickinson2015identifying}, researchers detect events such as `Falling in Love' and `Getting Married' by traditional machine learning models (e.g., SVM) with manually designed bag-of-words features. As pointed out by 
\citet{nguyen2017robust}, feature engineering based methods cannot effectively model word dependencies. To improve this, deep learning models on text are consequentially used, including CNN~\cite{nguyen2017robust,wu2019event} and Bi-LSTM~\cite{yen2018detecting,yen2019personal}. Comparing with social media text, key information in much longer news articles is inevitably mixed with and scattered around much more irrelevant information. Therefore, graph-based models with the ability to capture long-distance semantic dependency~\cite{yao2019graph}, may be a suitable choice for our task.


\subsection{Graph Neural Network}
Recently, graph neural network has been widely used in tasks such as text classification~\cite{alam2018graph,yao2019graph}, event prediction~\cite{deng2019learning}, and sentiment analysis~\cite{wang2020relational,li2021dual} due to its ability to capture long-range structural relationships on word graphs converted from the text. In our task, a major difference/difficulty is that news articles often contain trips of other celebrities, which distracts the model from focusing on the target celebrity. This can be improved, regarding the graph pooling technique applied to read out the graphs. The aforementioned methods rely on global pooling which cannot effectively distinguish key information (related to the target celebrity) from distractions~\cite{atwood2016diffusion,simonovsky2017dynamic} and this is later improved~\cite{lee2019self} by incorporating the attention mechanism that learns the importance automatically. In our scenario,
rather than letting the attention mechanism to decide the importance all by itself, we further guide it to focus on the target celebrity, to suit our ultimate task.

\section{Ethics Discussion and Broader Perspective}
Our study uses public news and Wikipedia data, and we ensure that our data collection process does not infringe upon any privacy or confidentiality concerns. A potential ethical concern is the misuse to track individual users (non-celebrities). However, as our framework relies on detailed text descriptions and background knowledge related to the person, the risk would stem from breaches of personal information rather than the framework itself. Therefore, we believe that our work does not raise essential ethical issues.
Regarding a broader perspective, our framework opens up the possibility to perform large-scale and network-wise studies on the travels and interactions of celebrities including politicians and artists, with implications in politics, economics, arts, and cultures.


\section{Conclusion and Future Work}

This paper proposes a new task of celebrity trip detection and introduces CeleTrip, a first framework that effectively extracts these trips from news articles, by modelling text and external knowledge as graphs. We construct the largest celebrity trip dataset known so far, on which our model outperforms other baselines. To facilitate relevant research on trip extraction as well as any analysis of celebrity trips, we open source the framework, the time and location extraction tools and the trip dataset. 
As a possible future step, we consider analyzing the patterns of the celebrity trips, given many historical trips, and extending the framework to various types of documents.

\bibliography{zreference_expand}

\end{document}